\pdfoutput=1
\documentclass[runningheads]{llncs}

\usepackage[T1]{fontenc}
\usepackage[utf8]{inputenc}
\usepackage[english]{babel}
\usepackage[dvipsnames,table]{xcolor}
\usepackage{amsmath}
\usepackage{amssymb}
\usepackage{booktabs}
\usepackage{xspace}
\usepackage{microtype}
\usepackage{graphicx}
\usepackage{subcaption}
\usepackage{url}
\usepackage{hyperref}
\usepackage{ifthen}
\hypersetup{
  colorlinks=true,
  linkcolor=blue,
  citecolor=ForestGreen,
  urlcolor=magenta
}
\newbool{anon}
\setboolean{anon}{false}
\usepackage[capitalise]{cleveref}
\title{TypeProbe: Recovering Type Representations from Hidden States of Pre-trained Code Models}
\titlerunning{TypeProbe: Recovering Type Representations}

\newcommand{\pyUnt}{\texttt{pyUnt}\xspace}
\newcommand{\pyTag}{\texttt{pyTag}\xspace}
\newcommand{\java}{\texttt{Java}\xspace}
\newcommand{\Java}{\texttt{Java}\xspace}

\newcommand{\githubLink}{\ifthenelse{\boolean{anon}}{\footnote{\url{https://anonymous.4open.science/r/TypeProbe-EA79/}}}{\footnote{\url{https://github.com/anticleiades/TypeProbe}}}}
\ifthenelse{\boolean{anon}}{
  \author{Anonymous Student Session Submission}
  \institute{Anonymous Institute}
  \authorrunning{Anonymous et al.}
}
{
  \author{Giuliano Gorgone \orcidID{0009-0004-0827-592X}, Fausto Carcassi \orcidID{0000-0001-6843-1737}}
  \authorrunning{G. Gorgone, F. Carcassi}
  \institute{ILLC, University of Amsterdam, Amsterdam, The Netherlands \\
  \email{giuliano.gorgone@student.uva.nl, f.carcassi@uva.nl}}
}

\begin{document}

\maketitle
\begin{abstract}
State-of-the-art code models achieve impressive performance, yet the extent to which they internally encode type information remains poorly understood. We probe the residual streams of pretrained code models for internal type representations using a parallel dataset of Java and Python code examples. Our results show that cross-lingual type representations emerge even from untyped code. Moreover, we test whether hidden states linearly encode the result type implied by typed function application by training probes on one language to infer argument and result types in the other. Finally, we find that this structure is partly robust to lexical perturbations and cross-language syntactic variations. To the best of our knowledge, prior work on interpretability of code models has not directly targeted formal type semantics or cross-lingual type representations. We release our code and datasets \githubLink.

\end{abstract}
\section{Introduction}
State-of-the-art code models excel at next-token prediction but can still violate formal type-system constraints. For instance, in generated TypeScript, nearly 94\% of compilation errors stem from type-check failures \cite{typeconstr}. Type-constrained decoding \cite{typeconstr} mitigates this by enforcing language-specific rules externally, but such approaches assume that models cannot internalize formal type information and require ad hoc machinery for each language.

Diagnostic probing has become a standard tool for studying abstract information encoded in neural representations \cite{AlainB17,hewitt-liang-2019-designing,belinkov-2022-probing}, and prior work on pretrained code models shows that they capture syntax, identifiers, and namespaces more readily than complex semantic properties \cite{troshin-chirkova-2022-probing}. At the same time, recent activation-steering results suggest that code models contain a controllable type-prediction mechanism shared across languages \cite{lucchetti2025}, motivating a more direct investigation of how such information is represented internally. In natural language, multilingual representations often exhibit approximately aligned manifolds across languages \cite{ammar2016massivelymultilingualwordembeddings,kudugunta2019investigating}; here we ask whether an analogous structure emerges for type information in code models.

In this work, we use layer-wise linear probes to test whether pretrained code models encode type information in their residual streams, where this information is localized across depth, and whether it transfers across Java and Python.
\section{Interpreting Code Models}
We investigate the internal logic of code models by addressing three primary research questions. First, we examine \textbf{representation}: do pretrained code models develop linearly decodable representations of type-level semantics, and in which transformer layers does this information emerge? Second, we explore \textbf{invariance and robustness}: is this representation invariant to language syntax and robust against lexical-level adversarial perturbations? More specifically, do models prioritize formal type constraints over superficial heuristics when identifier names explicitly contradict their true underlying types? Finally, we assess \textbf{inference}: beyond type prediction, can models leverage these internal representations to resolve data flow and perform typed function application?

\subsection{Experimental Setup}
\paragraph{Dataset and Design.}

We construct a programmatically generated dataset with three 90K-example partitions: \texttt{Java} (Java), \pyUnt (Python without type annotations), and \pyTag (Python with type annotations). All three partitions implement the same underlying programs, although the Java and Python versions differ in surface structure, since Java requires \texttt{class} and \texttt{main} declarations that Python omits. Program complexity ranges from simple one-line functions to more involved conditionals and loops. To prevent lexical shortcuts, identifiers and literals are independently randomized per partition, ensuring that operationally equivalent examples never share exact names or values across languages.

As shown in Figure~\ref{fig:fim_masking}, each example contains a masked call site of the form $b: \tau_{b} = \texttt{f}\_{\langle \mathbf{FIM} \rangle}(a)$, where $\langle \mathbf{FIM} \rangle$ denotes the fill-in-the-middle token. The model must recover the correct function using only type compatibility between $a$, $b$, and the candidate signatures. The fixed prefix \texttt{f\_} restricts prediction to a choice between the two candidate functions available in scope. Candidates always feature unique input types, enabling the task to be solved. We also uniformly distribute input-output type pairs to avoid trivial shortcuts.

\begin{figure}[t]
\centering
\setlength{\tabcolsep}{4pt}
\renewcommand{\arraystretch}{1.0}
\begin{tabular}{@{}p{0.46\linewidth}|p{0.46\linewidth}@{}}
\textbf{Python} & \textbf{Java} \\
\midrule
\begin{minipage}[t]{\linewidth}
\vspace{0pt}
\footnotesize
\begin{verbatim}
<fim-prefix>
def f_a(x):
    return x + 1

def f_b(xs):
    return len(xs)

n = 30328
y = f_<fim-suffix>(n)<fim-middle>
\end{verbatim}
\end{minipage}
&
\begin{minipage}[t]{\linewidth}
\vspace{0pt}
\footnotesize
\begin{verbatim}
<fim-prefix>
class Main {
  static int f_a(String s) {
    return s.length();
  }
  static int f_b(int x) {
    return x + 1;
  }
  static void main(String[] a) {
    int n = 666;
    int y = f_<fim-suffix>(n);
  }
}<fim-middle>
\end{verbatim}
\end{minipage}
\end{tabular}
\caption{Example dataset instances in Python (left) and Java (right) in SantaCoder's fill-in-the-middle (FIM) format. In both cases, the ground truth function is determined only by type compatibility at the call site: \texttt{f\_a} in Python and \texttt{f\_b} in Java.}
\label{fig:fim_masking}
\end{figure}
\paragraph{Type Systems.}
We focus on operationally equivalent types across Python ($\mathcal{T}_P$) and Java ($\mathcal{T}_J$) by establishing a direct mapping between base types (\texttt{str} $\leftrightarrow$ \texttt{String}, \texttt{bool} $\leftrightarrow$ \texttt{boolean}, \texttt{int} $\leftrightarrow$ \texttt{int}, \texttt{float} $\leftrightarrow$ \texttt{float}) and their respective \texttt{list} containers. Note that Python lists and Java lists are represented by different token sequences (e.g., \texttt{["a", "b"]} vs.\ \texttt{Arrays.asList("a", "b")}), and even their type annotations differ (\texttt{list[T]} in Python vs.\ \texttt{List<T>} in Java).

\paragraph{Adversarial Examples.}
To disentangle lexical cues from type-level reasoning, we construct an adversarial partition in which identifiers systematically contradict their true types. For every variable $v$ with ground-truth type $\tau \in \mathcal{T}$, we change its identifier to \texttt{var\_\{$\tau'$\}}, where $\tau' \in \mathcal{T} \setminus \{\tau\}$ is sampled uniformly from the set of incorrect types; similarly, function names are reassigned to \texttt{f\_\{$\tau_r'$\}}, where the identifier intentionally contradicts the true return type $\tau_r$. In the following, we refer to the original data as the \emph{standard} partition.

\paragraph{Models.}
We evaluate two decoder-only Transformer code models: SantaCoder-1.1B (24 layers, 16 heads, hidden size 2048, MQA) \cite{allal2023santacoderdontreachstars} and CodeLlama-7B (32 layers, 32 heads, hidden size 4096, MHA) \cite{roziere2024codellamaopenfoundation}.

\paragraph{Probing Tasks.}
We define three tasks based on the masked site $b : \tau_{b} = \texttt{f}\_{\langle \mathbf{FIM} \rangle}(a)$, where $a$ has type $\tau_a$:
\textbf{(Task 0) Function Prediction:} binary classification of the invoked candidate;
\textbf{(Task 1) Argument Type:} 8-class prediction of $\tau_a$;
\textbf{(Task 2) Application Result:} 8-class prediction of the return type of the function, which is $\tau_b$. This task allows us to probe a simple form of compositional type semantics, namely whether the hidden states encode the result of a basic yet non-trivial inferential step in typed function application \footnote{Here, ``inferential step'' refers to the process of inferring function return type based on its signature and the provided input type. We use this term solely to describe the structure recoverable by the probe, not to claim general reasoning abilities.}. Disentangling these tasks allows us to isolate failure modes, such as successful result-type prediction (Task 2) despite incorrect function selection (Task 0).

\paragraph{Probing Setup.}
For each input, we extract the post-residual activation at the FIM token position, i.e., the step at which the model predicts the masked function from the preceding context and argument. We exhaustively probe all layers by training a separate linear probe for each model, layer, task, training partition, and evaluation setting. Probes are trained with cross-entropy loss using AdamW (learning rate $10^{-4}$, weight decay $0.01$) for 20 epochs, without feature normalization. We use 4-fold cross-validation, holding out $10\%$ of each training fold for in-distribution validation; for each task, the best checkpoint is selected by validation accuracy and evaluated on the test fold. Batch size is 512 for CodeLlama and 256 for SantaCoder. All experiments were run on a single NVIDIA H100 GPU (80\,GB) using PyTorch and TransformerLens \cite{nanda2022transformerlens}.

\paragraph{Evaluation Metrics and Methodology.}
We measure and compare across models:
\textbf{(1) Selectivity:} following \cite{hewitt-liang-2019-designing}, we define selectivity as $S = \text{acc}_{\text{task}} - \text{acc}_{\text{control}}$, where $\text{acc}_{\text{control}}$ is the performance of an architecturally identical probe trained on a \emph{control task} constructed \emph{per-token}, i.e., by assigning random labels within the same output space as the original task. While \cite{hewitt-liang-2019-designing} uses a \emph{per-type} control, we adopt a \emph{per-token} control, which we consider a reasonable approximation under our data construction, since identifiers and literals are independently randomized per example and across partitions. \textbf{(2) Robustness:} we quantify \emph{robustness} via absolute drop in \emph{selectivity} $\Delta$ between the \emph{standard} and the \emph{adversarial} setting. Furthermore, we evaluate cross-language generalization through zero-shot transfer by training on one partition (\Java, \pyUnt, or \pyTag) and testing on the remaining out-of-distribution partitions to assess whether type information transfers across languages.

\subsection{Results and Discussion}
\begin{table*}[t]
    \centering
    \caption{\textbf{Task 0: Function Prediction.} Values report peak selectivity $\mathcal{S}$ ($\times 100$) and peak layer ($L$). Bold indicates values corresponding to at least $75\%$ raw accuracy ($S \ge 25$); italics indicate non-negligible signals ($S \ge 20$).}
    \label{tab:task0-results}
    \small
    \setlength{\tabcolsep}{4pt}
    \resizebox{\textwidth}{!}{
    \begin{tabular}{@{} ll ccc ccc ccc @{} }
        \toprule
        & & \multicolumn{3}{c}{\textbf{Eval on Standard}} & \multicolumn{3}{c}{\textbf{Eval on Adversarial}} & \multicolumn{3}{c}{\textbf{Absolute Drop (\boldmath$\Delta$)}} \\
        \cmidrule(lr){3-5} \cmidrule(lr){6-8} \cmidrule(lr){9-11}
        \textbf{Model} & \textbf{Train} & \textbf{Java} & \textbf{pyTag} & \textbf{pyUnt} & \textbf{Java} & \textbf{pyTag} & \textbf{pyUnt} & \textbf{Java} & \textbf{pyTag} & \textbf{pyUnt} \\
        \midrule
        \textbf{CodeLlama (7B)} & Java & \textbf{\textit{50}} (16) & \textbf{\textit{42}} (16) & 04 (16) & \textbf{\textit{43}} & \textbf{\textit{37}} & 02 & 06 & 05 & 01 \\
         & pyTag & \textbf{\textit{47}} (17) & \textbf{\textit{47}} (17) & 07 (17) & \textbf{\textit{44}} & \textbf{\textit{40}} & 03 & 02 & 08 & 04 \\
         & pyUnt & \textbf{\textit{29}} (13) & \textbf{\textit{29}} (13) & 12 (13) & \textit{23} & \textit{20} & 04 & 06 & 08 & 08 \\
        \midrule
        \textbf{SantaCoder (1.1B)} & Java & \textbf{\textit{47}} (13) & \textbf{\textit{38}} (13) & 01 (13) & \textbf{\textit{37}} & \textbf{\textit{27}} & 00 & 10 & 11 & 02 \\
         & pyTag & \textbf{\textit{44}} (13) & \textbf{\textit{43}} (13) & 02 (13) & \textbf{\textit{37}} & \textbf{\textit{25}} & -1 & 07 & 18 & 03 \\
         & pyUnt & 16 (12) & \textbf{\textit{25}} (12) & 04 (12) & 16 & \textbf{\textit{28}} & 01 & 00 & -3 & 03 \\
        \bottomrule
    \end{tabular}}
\end{table*}

\begin{table*}[t]
  \centering
  \caption{\textbf{Task 1: Argument Type Prediction.} Values report peak selectivity $\mathcal{S}$ ($\times 100$) and peak layer ($L$). Bold indicates values corresponding to at least $75\%$ raw accuracy ($S \ge 63$); italics indicate non-negligible signals ($S \ge 30$).}
  \label{tab:task1-results}
  \small
  \setlength{\tabcolsep}{4pt}
  \resizebox{\textwidth}{!}{
    \begin{tabular}{@{} ll ccc ccc ccc @{} }
      \toprule
      & & \multicolumn{3}{c}{\textbf{Eval on Standard}} & \multicolumn{3}{c}{\textbf{Eval on Adversarial}} & \multicolumn{3}{c}{\textbf{Absolute Drop (\boldmath$\Delta$)}} \\
      \cmidrule(lr){3-5} \cmidrule(lr){6-8} \cmidrule(lr){9-11}
      \textbf{Model} & \textbf{Train} & \textbf{Java} & \textbf{pyTag} & \textbf{pyUnt} & \textbf{Java} & \textbf{pyTag} & \textbf{pyUnt} & \textbf{Java} & \textbf{pyTag} & \textbf{pyUnt} \\
      \midrule
      \textbf{CodeLlama (7B)} & Java & \textbf{\textit{87}} (16) & \textbf{\textit{77}} (16) & \textbf{\textit{73}} (16) & \textbf{\textit{74}} & \textit{36} & \textit{39} & 12 & 41 & 33 \\
      & pyTag & \textit{61} (16) & \textbf{\textit{86}} (16) & \textbf{\textit{82}} (16) & \textit{41} & \textbf{\textit{64}} & \textit{58} & 19 & 21 & 24 \\
      & pyUnt & \textit{47} (16) & \textbf{\textit{83}} (16) & \textbf{\textit{84}} (16) & 28 & \textit{48} & \textit{61} & 19 & 35 & 23 \\
      \midrule
      \textbf{SantaCoder (1.1B)} & Java & \textbf{\textit{87}} (13) & \textbf{\textit{68}} (13) & \textbf{\textit{64}} (13) & \textbf{\textit{71}} & \textit{51} & \textit{53} & 16 & 17 & 11 \\
      & pyTag & \textit{48} (11) & \textbf{\textit{86}} (11) & \textbf{\textit{73}} (11) & 27 & \textbf{\textit{70}} & \textit{48} & 21 & 17 & 25 \\
      & pyUnt & \textbf{\textit{73}} (11) & \textbf{\textit{86}} (11) & \textbf{\textit{85}} (11) & \textit{53} & \textbf{\textit{77}} & \textbf{\textit{76}} & 20 & 09 & 09 \\
      \bottomrule
  \end{tabular}}
\end{table*}

\begin{table*}[t]
  \centering
  \caption{\textbf{Task 2: Application Result Prediction.} Values report peak selectivity $\mathcal{S}$ ($\times 100$) and peak layer ($L$). Formatting and thresholds follows Table~\ref{tab:task1-results}.}
  \label{tab:task2-results}
  \small
  \setlength{\tabcolsep}{4pt}
  \resizebox{\textwidth}{!}{
    \begin{tabular}{@{} ll ccc ccc ccc @{} }
      \toprule
      & & \multicolumn{3}{c}{\textbf{Eval on Standard}} & \multicolumn{3}{c}{\textbf{Eval on Adversarial}} & \multicolumn{3}{c}{\textbf{Absolute Drop (\boldmath$\Delta$)}} \\
      \cmidrule(lr){3-5} \cmidrule(lr){6-8} \cmidrule(lr){9-11}
      \textbf{Model} & \textbf{Train} & \textbf{Java} & \textbf{pyTag} & \textbf{pyUnt} & \textbf{Java} & \textbf{pyTag} & \textbf{pyUnt} & \textbf{Java} & \textbf{pyTag} & \textbf{pyUnt} \\
      \midrule
      \textbf{CodeLlama (7B)} & Java & \textbf{\textit{91}} (14) & \textit{45} (14) & 04 (14) & \textbf{\textit{80}} & 25 & -4 & 11 & 20 & 08 \\
      & pyTag & \textbf{\textit{84}} (17) & \textbf{\textit{83}} (17) & 20 (17) & \textbf{\textit{77}} & \textbf{\textit{69}} & -3 & 07 & 14 & 23 \\
      & pyUnt & \textit{45} (18) & \textbf{\textit{74}} (18) & \textit{44} (18) & 25 & \textit{55} & \textit{32} & 20 & 20 & 11 \\
      \midrule
      \textbf{SantaCoder (1.1B)} & Java & \textbf{\textit{83}} (12) & \textbf{\textit{70}} (12) & 15 (12) & \textbf{\textit{72}} & \textit{54} & 02 & 11 & 16 & 13 \\
      & pyTag & \textbf{\textit{77}} (12) & \textbf{\textit{79}} (12) & 17 (12) & \textit{53} & \textbf{\textit{68}} & -2 & 25 & 12 & 19 \\
      & pyUnt & \textit{53} (12) & \textit{54} (12) & \textit{40} (12) & 27 & \textit{39} & \textit{32} & 27 & 15 & 08 \\
      \bottomrule
  \end{tabular}}
\end{table*}

Tables~\ref{tab:task0-results}--\ref{tab:task2-results} report selectivity $S$ ($\times 100$) and peak layer $L$ for each model and training partition under \emph{standard} and \emph{adversarial} evaluation. Peak layers are selected on a held-out validation split, and adversarial scores are measured at the same layer selected in the standard setting. The final columns report the absolute selectivity drop $\Delta$ from standard to adversarial. Following prior work \cite{AlainB17,belinkov-2022-probing}, high selectivity indicates linearly decodable task information, while low selectivity suggests little structure beyond the randomized-label control \cite{belinkov-2022-probing}. Since $\text{acc}_{\text{control}}$ consistently approximates chance level, selectivity $\mathcal{S} \in [-50.0, 50.0]$ for Task~0 and $\mathcal{S} \in [-12.5, 87.5]$ for Task 1 and Task 2.

\paragraph{Cross-Lingual Type Representation.}
Results in \cref{tab:task1-results,tab:task2-results} show that both \textbf{SantaCoder-1.1B} and \textbf{CodeLlama-7B}\footnote{We thank the reviewers for highlighting tokenization as a potential confounding factor. We then identified a preprocessing issue in our initial CodeLlama setup: the FIM marker was being split into four separate tokens rather than processed as a single special token. Correcting this improved CodeLlama's probing accuracy, and all reported results now reflect this corrected setup.} develop linearly decodable type representations that transfer across languages.

\begin{figure}[h]
  \centering
  \includegraphics[width=\linewidth]{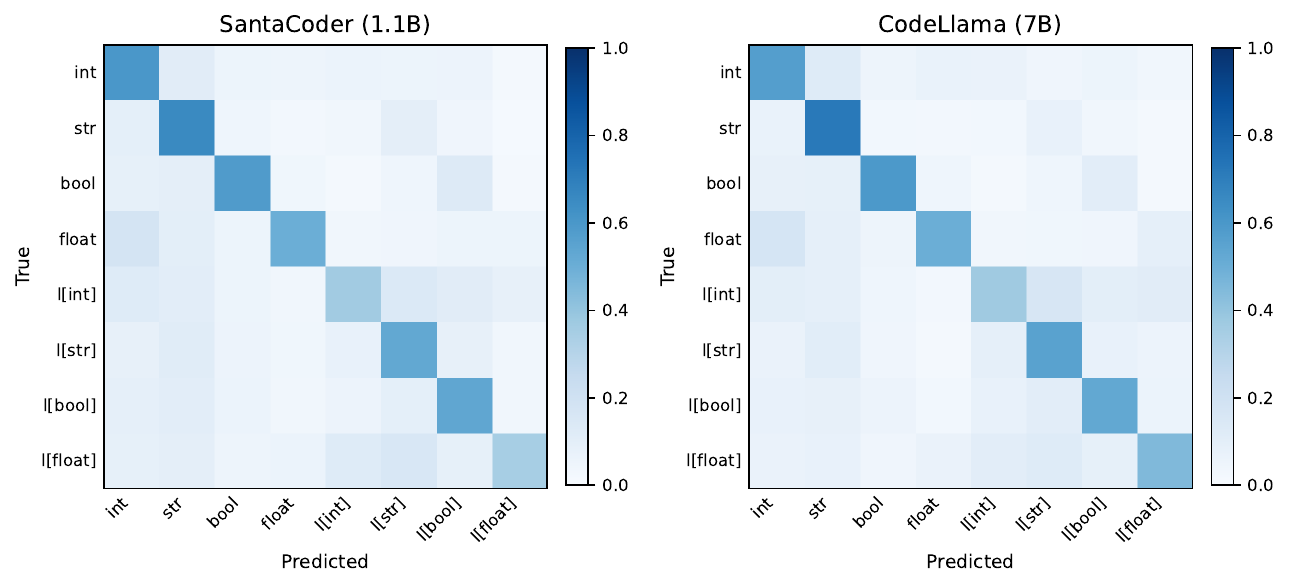}
  \vspace{0.2cm} 
  \includegraphics[width=\linewidth]{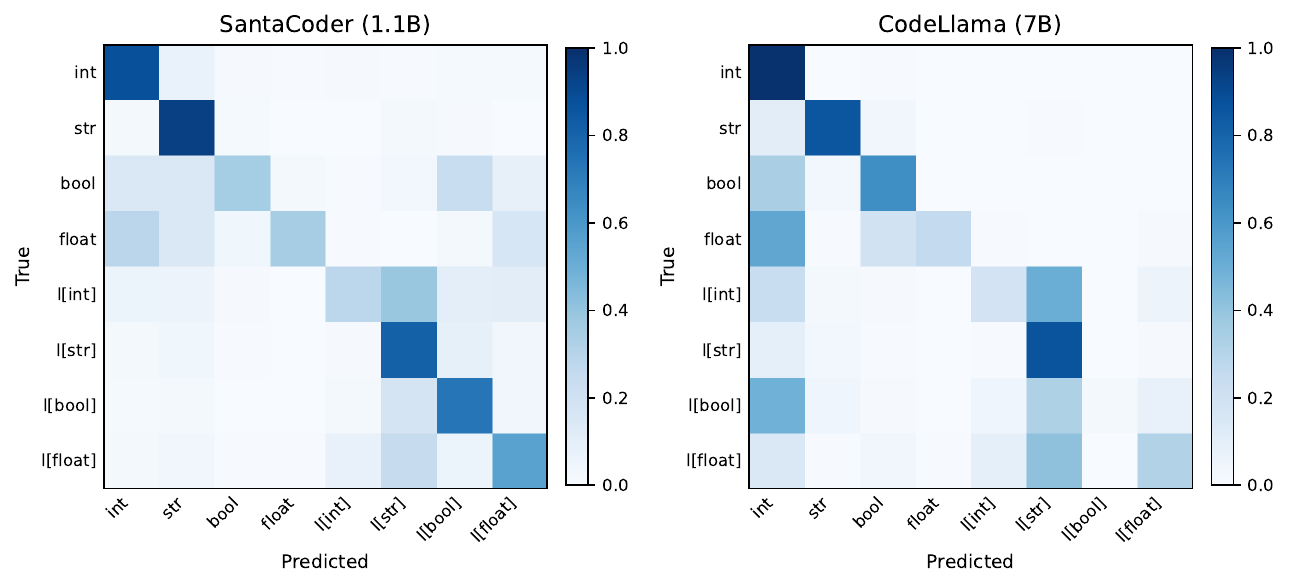}
  \caption{Row-normalized confusion matrices for Task 2 (Application Result) with probes trained on \pyUnt. \textbf{Top:} Baseline (evaluated on \pyUnt). \textbf{Bottom:} Zero-shot transfer (evaluated on \Java). In both panels, SantaCoder (layer 12) is on the left and CodeLlama (layer 18) is on the right.}
  \label{fig:cm-task2}
\end{figure}

\textbf{SantaCoder} exhibits a more uniform cross-lingual transfer profile than CodeLlama. In Task~1, \Java-trained probes transfer well to \texttt{pyTag} (\(68\)) and \texttt{pyUnt} (\(64\)), while \texttt{pyUnt}-trained probes generalize back to \Java (\(73\)). In Task~2, transfer remains robust between \Java and \texttt{pyTag} (\(70\)--\(77\)) and remains meaningful for \texttt{pyUnt} \(\rightarrow\) \Java (\(53\)). The confusion matrices in \cref{fig:cm-task2} further support this interpretation: under cross-lingual transfer, SantaCoder preserves separability for most base and list types, with only modest increases in confusion among numeric types and among list types.

Regarding \textbf{CodeLlama}, in Task~1, transfer between \Java and \texttt{pyUnt} remains non-trivial in both directions, with \Java \(\rightarrow\) \texttt{pyUnt} reaching \(73\) and \texttt{pyUnt} \(\rightarrow\) \Java reaching \(47\). In Task~2, by contrast, transfer is much less uniform: \texttt{pyTag} \(\rightarrow\) \Java reaches \(84\), \texttt{pyUnt} \(\rightarrow\) \Java reaches \(45\), but \Java \(\rightarrow\) \texttt{pyUnt} falls to \(04\). Unlike SantaCoder, CodeLlama also tends to over-predict \texttt{int} under cross-lingual transfer (\cref{fig:cm-task2}).

Crucially, transfer from \texttt{pyUnt} to \Java is particularly informative for both models. Because \texttt{pyUnt} lacks explicit annotations, successful transfer suggests that the probes recover type-relevant directions from untyped code alone. When explicit type information is present in the target activations (\Java and \pyTag), it appears to align with the same type-relevant signal already recoverable from untyped code. Task~0 provides additional support for this interpretation, showing non-trivial cross-lingual transfer in function prediction even when the probe is trained on examples without explicit type annotations.

\begin{figure}[htbp]
  \centering
  \includegraphics[width=\linewidth]{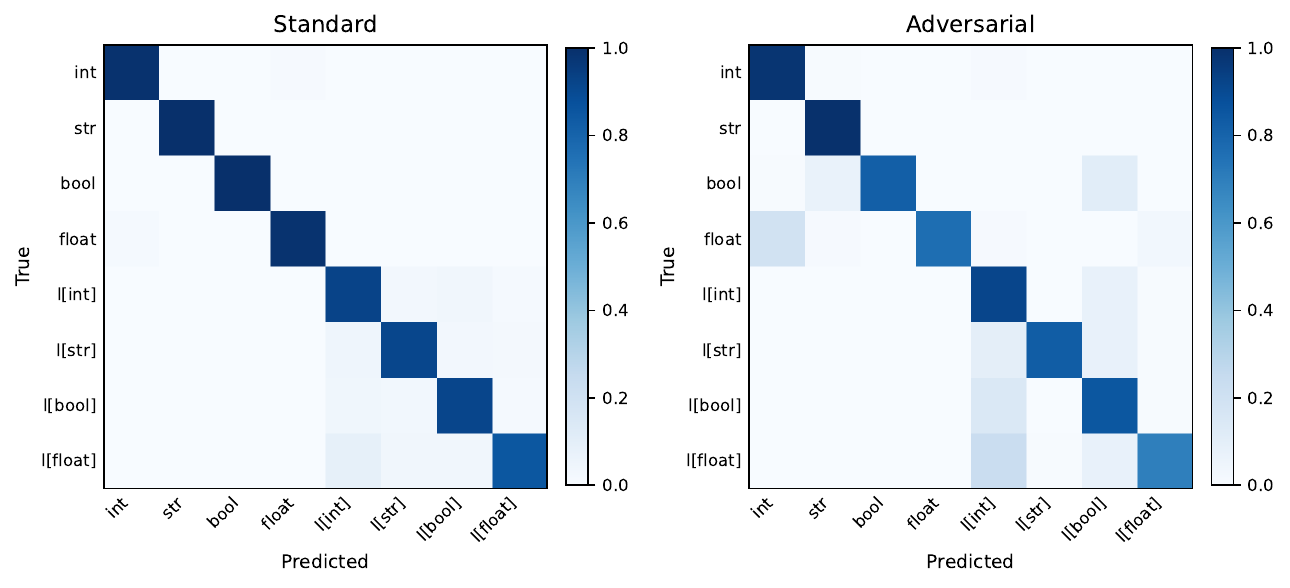}
  \caption{Row-normalized confusion matrices for \textbf{SantaCoder (1.1B)} at layer 11 on Task 1. Probes are trained and evaluated on \pyUnt\ under standard (left) and adversarial (right) settings.}
  \label{fig:cm-santacoder-task1}
\end{figure}

\begin{figure}[htbp]
  \centering
  \includegraphics[width=\linewidth]{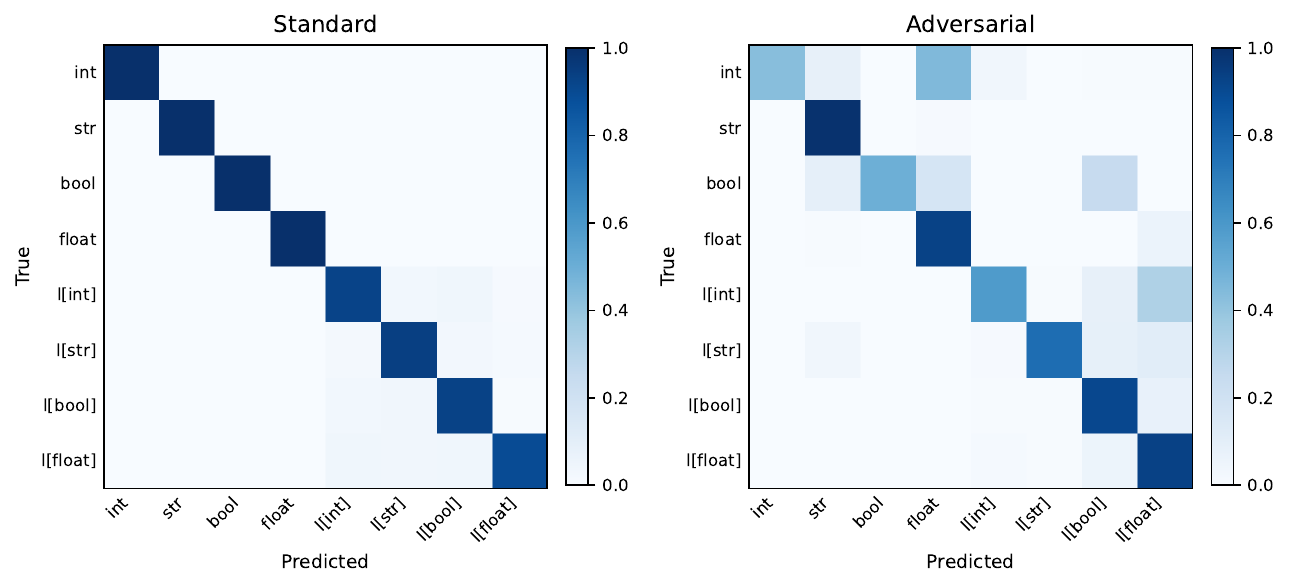}
  \caption{Row-normalized confusion matrices for \textbf{CodeLlama (7B)} at layer 16 on Task 1. Probes are trained and evaluated on \pyUnt\ under standard (left) and adversarial (right) settings.}
  \label{fig:cm-codellama-task1}
\end{figure}

In both models, the confusion matrices for \texttt{pyUnt} suggest that the probes often recover container structure more reliably than the element type (\cref{fig:cm-codellama-task1,fig:cm-santacoder-task1,fig:cm-task2,fig:cm-task2-pyUnt-combined}). Under the standard setting, this pattern appears less pronounced in in-domain \pyTag and \Java evaluation, where the confusion matrices are nearly diagonal (\cref{fig:cm-task1-pyTag-combined,fig:cm-task1-java-combined,fig:cm-task2-pyTag-combined,fig:cm-task2-java-combined}).

\begin{figure}[!ht]
  \centering
  \includegraphics[width=\textwidth]{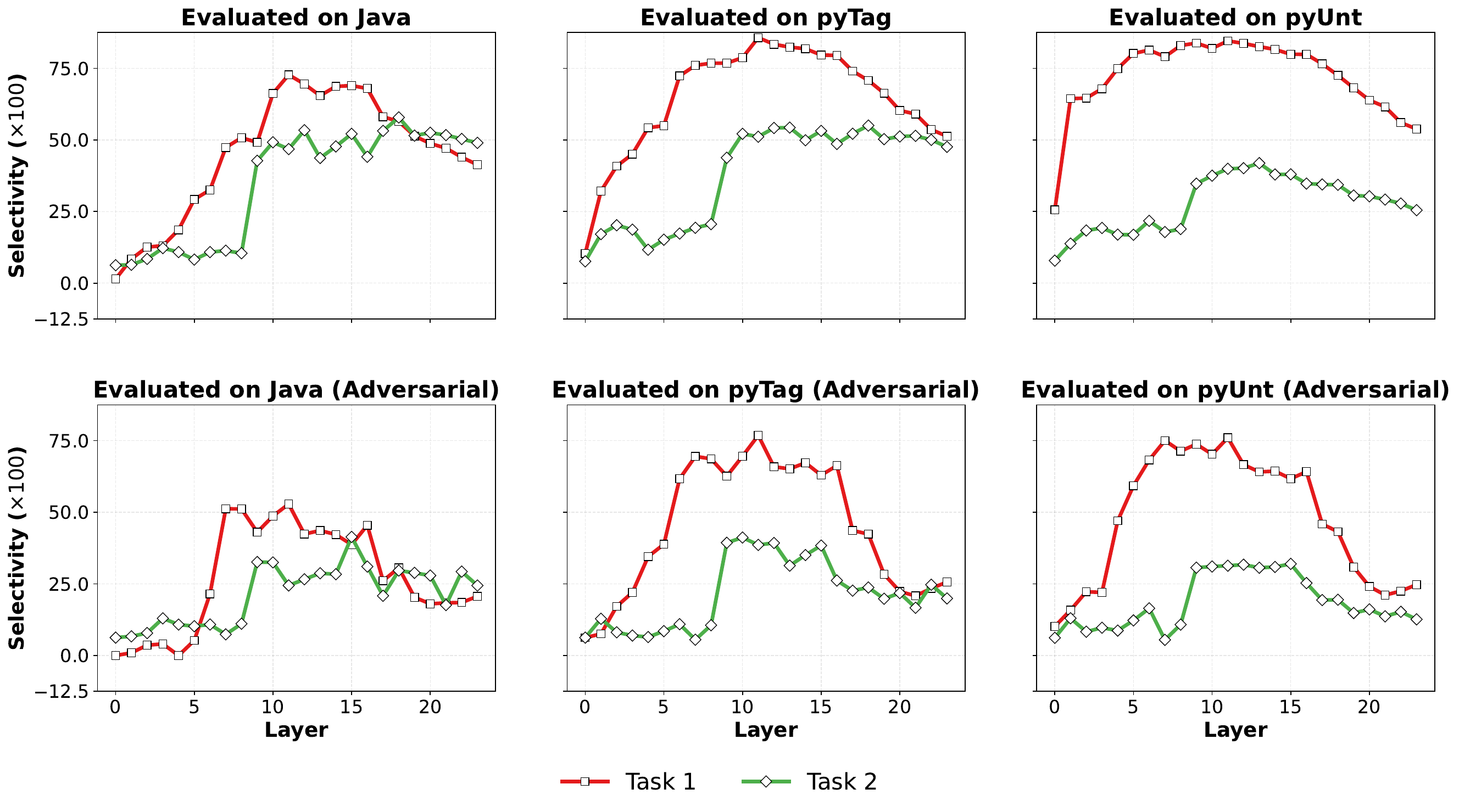}
  \vspace{0.2cm}
  \includegraphics[width=\textwidth]{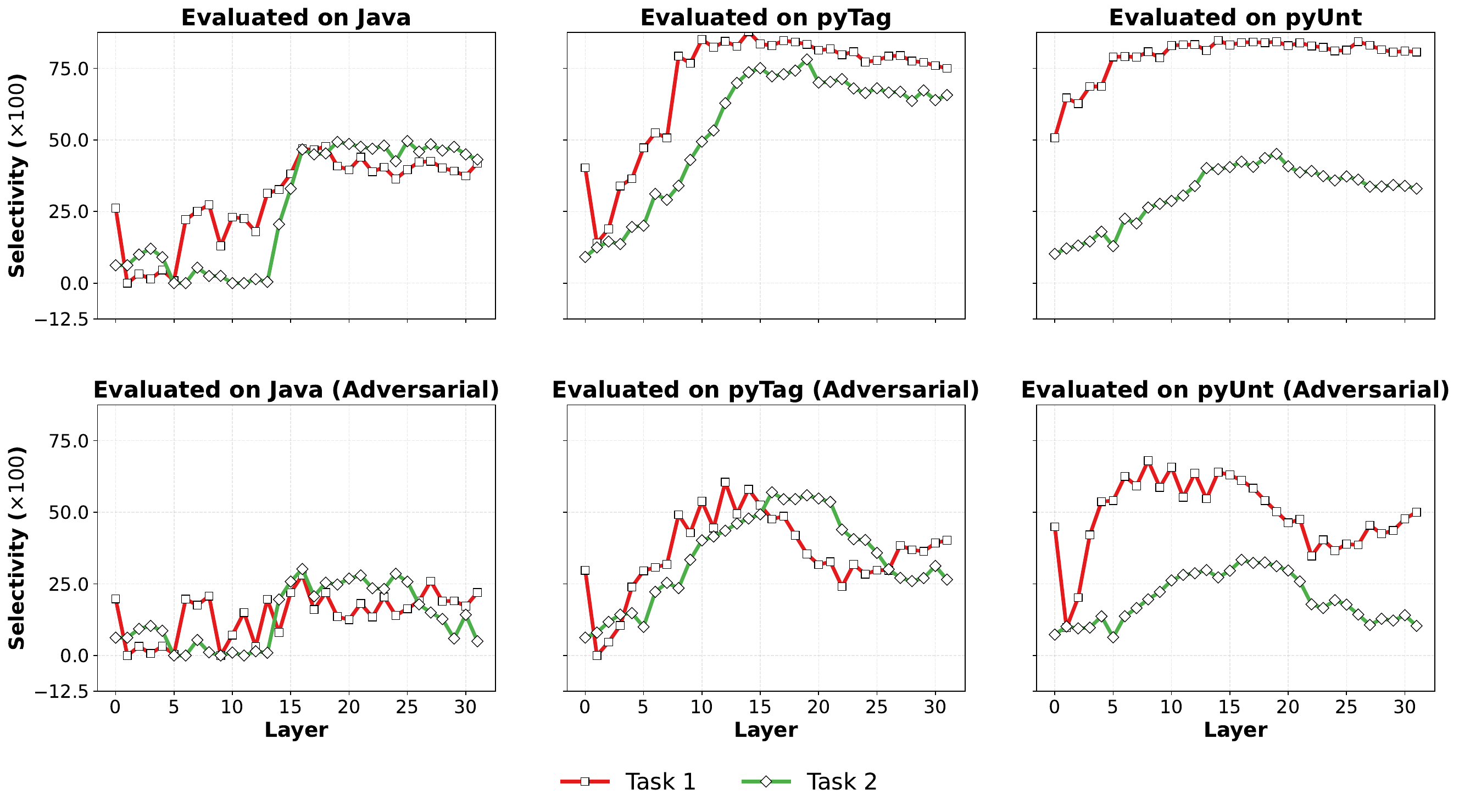}
  \caption{Layer-wise selectivity and accuracy (Tasks 1 \& 2) with probes trained on \texttt{pyUnt} and evaluated across all standard and adversarial partitions. \textbf{Top:} SantaCoder (1.1B). \textbf{Bottom:} CodeLlama (7B).}
  \label{fig:selectivity-combined}
\end{figure}

\paragraph{Per-layer Probing Performance.}
Selectivity peaks in mid-to-late layers (SantaCoder: L11--L16, CodeLlama: L14--L18; \cref{fig:selectivity-combined}). Notably, \pyUnt-trained probes exhibit a late-layer decodability drop that amplifies under adversarial evaluation. A similar pattern arises for explicitly typed partitions (\Java and \pyTag): while late-layer selectivity is relatively stable in standard settings, it consistently drops across all models when facing adversarial renaming (\cref{fig:selectivity-combined-java,fig:selectivity-combined-pyTag}).

\paragraph{Disentangling Type-Level Semantics from Lexical-Level Semantics.}
Figures~\ref{fig:cm-santacoder-task1}--\ref{fig:cm-codellama-task1} quantify the effect of adversarial lexical interference on type information decodability. Adversarial renaming shifts SantaCoder toward \texttt{int} \(\rightarrow\) \texttt{float} confusion, while CodeLlama shows the reverse pattern, with stronger \texttt{float} \(\rightarrow\) \texttt{int} errors. In both models, \texttt{str} remains comparatively stable, whereas the clustering among list types becomes more pronounced than in the baseline.
Qualitatively similar patterns of confusion also appear in \pyTag and \Java under adversarial evaluation, indicating that misleading identifier names can interfere with the decodability of type information even when explicit type information is available (\cref{fig:cm-task1-pyTag-combined,fig:cm-task1-java-combined,fig:cm-task2-pyTag-combined,fig:cm-task2-java-combined}).

\section{Conclusions}
Using linear probes \cite{belinkov-2022-probing}, we showed that SantaCoder and CodeLlama develop linearly decodable type representations that transfer across programming languages. These representations emerge even when probes are trained on untyped Python, suggesting that code models encode non-trivial type-relevant structure beyond explicit annotations. Across tasks, adversarial renaming reveals only partial robustness to misleading lexical cues. Overall, our findings point to the emergence of a cross-lingual type manifold in the setting we study.

\section{Limitations and Future Work}
\label{sec:limitations}
Our experimental setup focuses on base types and lists across two languages and models. Future work should explore richer relations (e.g., subtyping, covariance \cite{understandingTypes}) and lower-level type systems (e.g., Rust's ownership, C++'s templates) across varying model scales. Methodologically, while linear probes confirm \emph{decodability}, they cannot recover features in superposition \cite{elhage2022superposition,belinkov-2022-probing}, nor prove any causal role of the extracted representations in code generation. Future work should therefore combine our setup with more expressive tools, such as \emph{Sparse Autoencoders (SAEs)}, \emph{Dictionary Learning}, and representation-similarity methods like SVCCA \cite{kudugunta2019investigating}, as well as causal analyses via activation patching \cite{zhang2024towards}.
Ultimately, our findings suggest that learned type representations may complement type-constrained decoding approaches \cite{typeconstr}.

\begin{credits}
  \subsubsection{\ackname}
  \ifthenelse{\boolean{anon}}{
    Acknowledgments removed for blind review.
  }
  {We acknowledge SURF for computing access on Snellius and the ILLC (University of Amsterdam) for travel support. We thank Michael Hanna, Francesca Lucchetti and the anonymous reviewers for their insightful feedback, Adhvaith Koduru and Songyun Zou for their contributions to the preliminary experiments, and ESSLLI for a registration fee waiver.}
  \subsubsection{\discintname}
  \ifthenelse{\boolean{anon}}{
    No competing interests to declare that are relevant to the content of this article.
  }{ No competing interests to declare that are relevant to the content of this article.}
\end{credits}
\clearpage
\bibliographystyle{splncs04}
\bibliography{bib}
\clearpage 
\appendix
\section{Confusion Matrices: Task1 (\Java, \pyTag)}

\begin{figure}[h]
  \centering
  \begin{subfigure}{\linewidth}
    \centering
    \includegraphics[width=\linewidth]{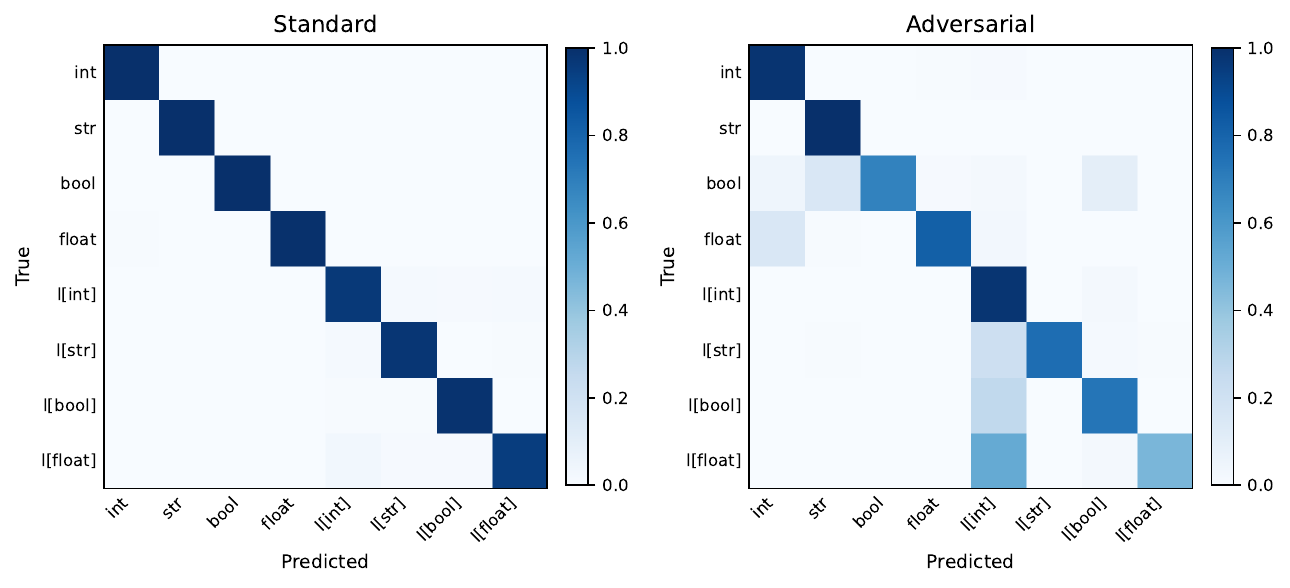}
    \caption{SantaCoder (1.1B) at layer 11.}
    \label{fig:cm-santacoder-task1-pyTag}
  \end{subfigure}
  
  \vspace{1.5em}
  
  \begin{subfigure}{\linewidth}
    \centering
    \includegraphics[width=\linewidth]{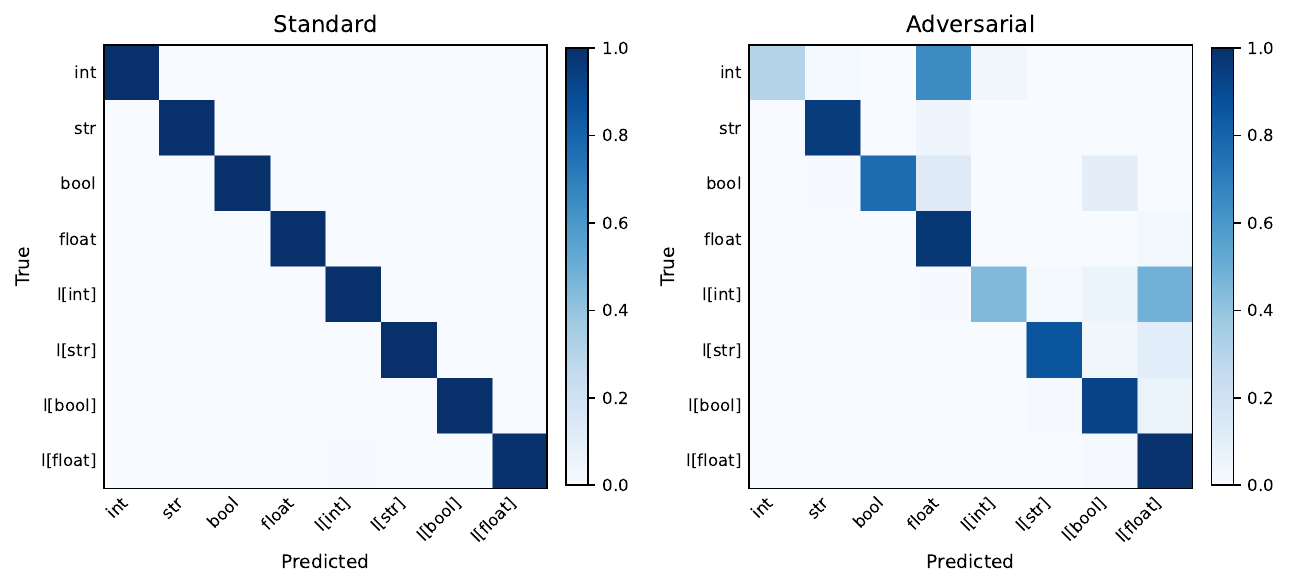}
    \caption{CodeLlama (7B) at layer 16.}
    \label{fig:cm-codellama-task1-pyTag}
  \end{subfigure}

  \caption{Row-normalized confusion matrices on Task 1. Probes are trained and evaluated on \pyTag\ under standard (left) and adversarial (right) settings.}
  \label{fig:cm-task1-pyTag-combined}
\end{figure}

\begin{figure}[h]
  \centering
  \begin{subfigure}{\linewidth}
    \centering
    \includegraphics[width=\linewidth]{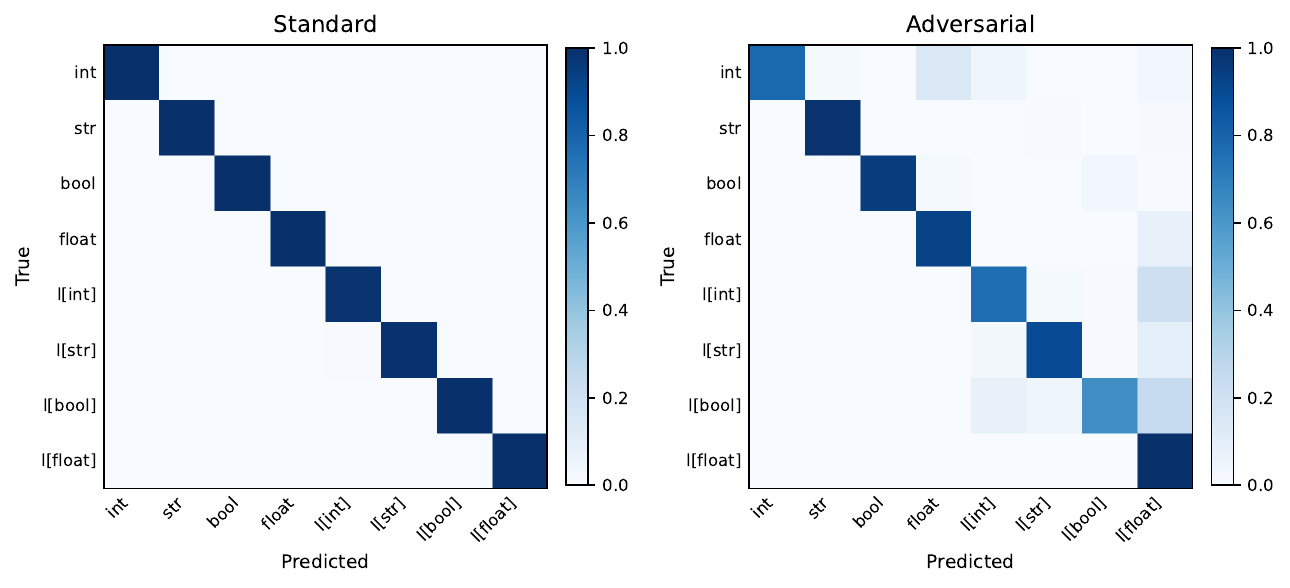}
    \caption{SantaCoder (1.1B) at layer 13.}
    \label{fig:cm-santacoder-task1-java}
  \end{subfigure}
  
  \vspace{1.5em}
  
  \begin{subfigure}{\linewidth}
    \centering
    \includegraphics[width=\linewidth]{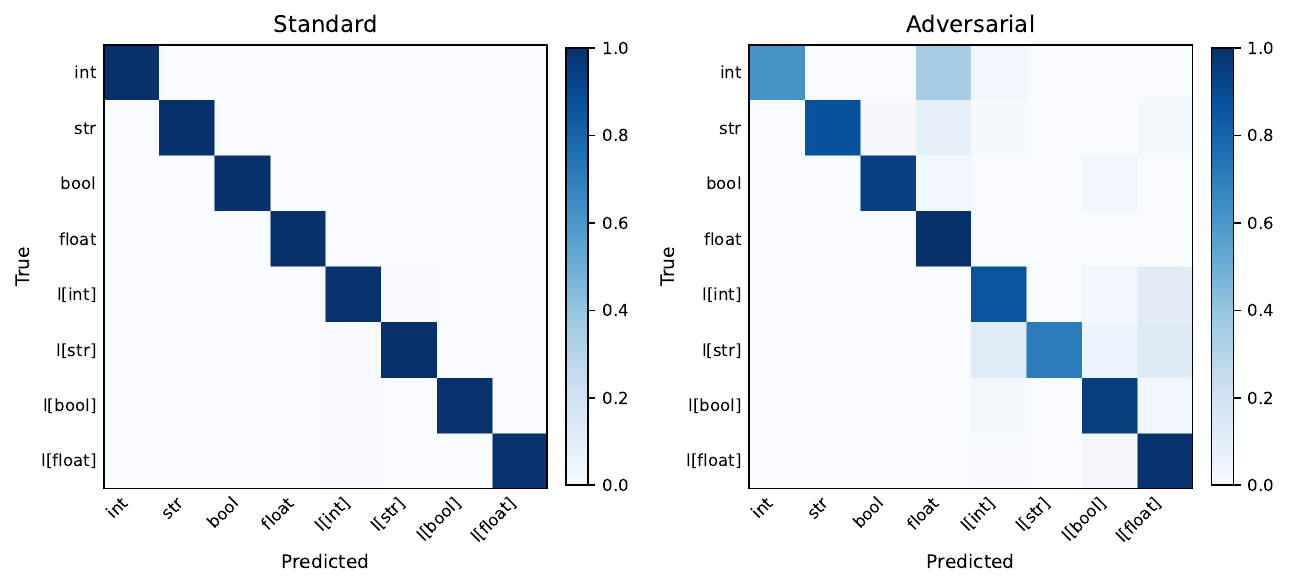}
    \caption{CodeLlama (7B) at layer 16.}
    \label{fig:cm-codellama-task1-java}
  \end{subfigure}

  \caption{Row-normalized confusion matrices on Task 1. Probes are trained and evaluated on \java under standard (left) and adversarial (right) settings.}
  \label{fig:cm-task1-java-combined}
\end{figure}
\clearpage

\section{Confusion Matrices: Task2 (\Java, \pyTag, \pyUnt)}

\begin{figure}[h]
  \centering
  \begin{subfigure}{\linewidth}
    \centering
    \includegraphics[width=\linewidth]{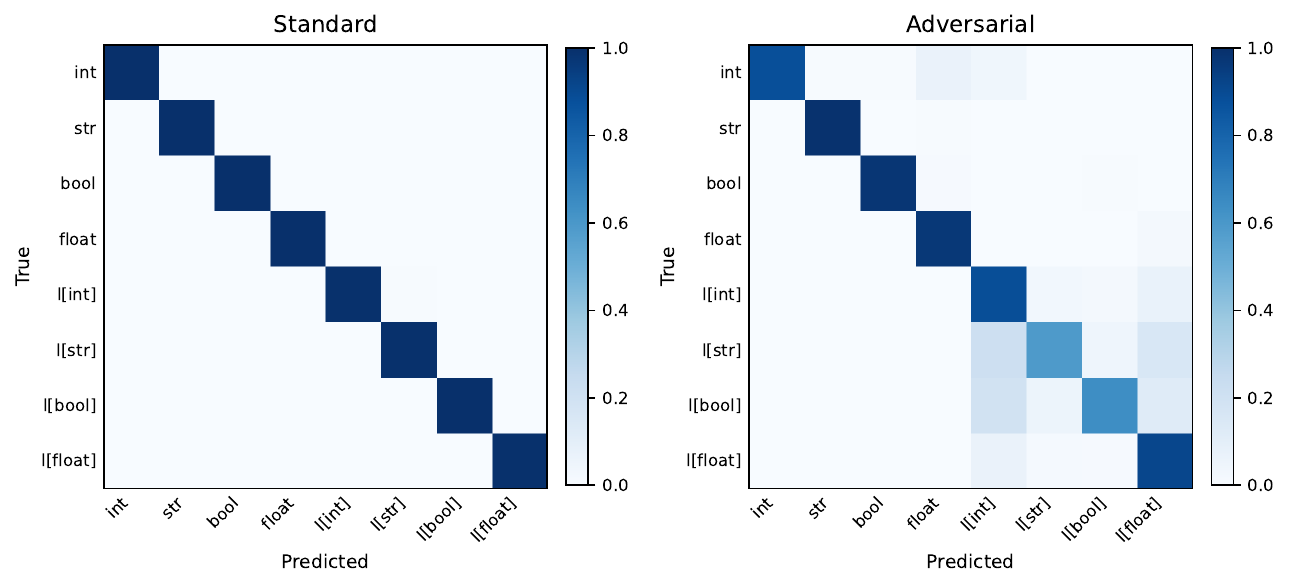}
    \caption{SantaCoder (1.1B) at layer 12.}
    \label{fig:cm-santacoder-task2-java}
  \end{subfigure}
  
  \vspace{1.5em}
  
  \begin{subfigure}{\linewidth}
    \centering
    \includegraphics[width=\linewidth]{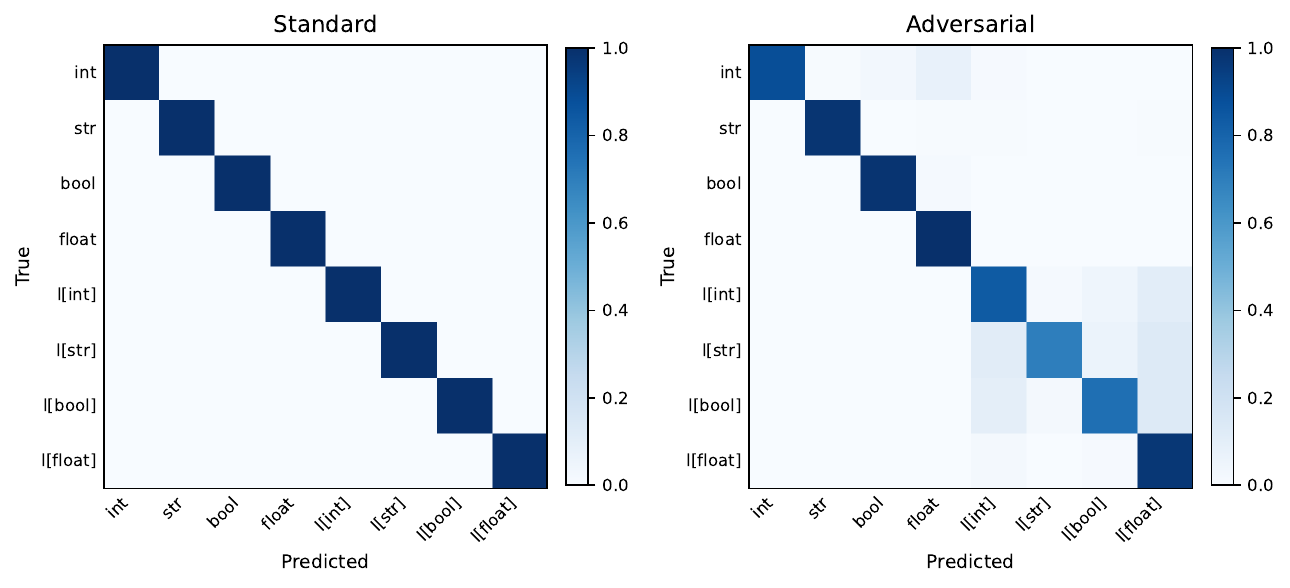}
    \caption{CodeLlama (7B) at layer 14.}
    \label{fig:cm-codellama-task2-java}
  \end{subfigure}

  \caption{Row-normalized confusion matrices for Task 2 (Application Result). Probes are trained and evaluated on \Java\ under standard (left) and adversarial (right) settings.}
  \label{fig:cm-task2-java-combined}
\end{figure}

\begin{figure}[h]
  \centering
  \begin{subfigure}{\linewidth}
    \centering
    \includegraphics[width=\linewidth]{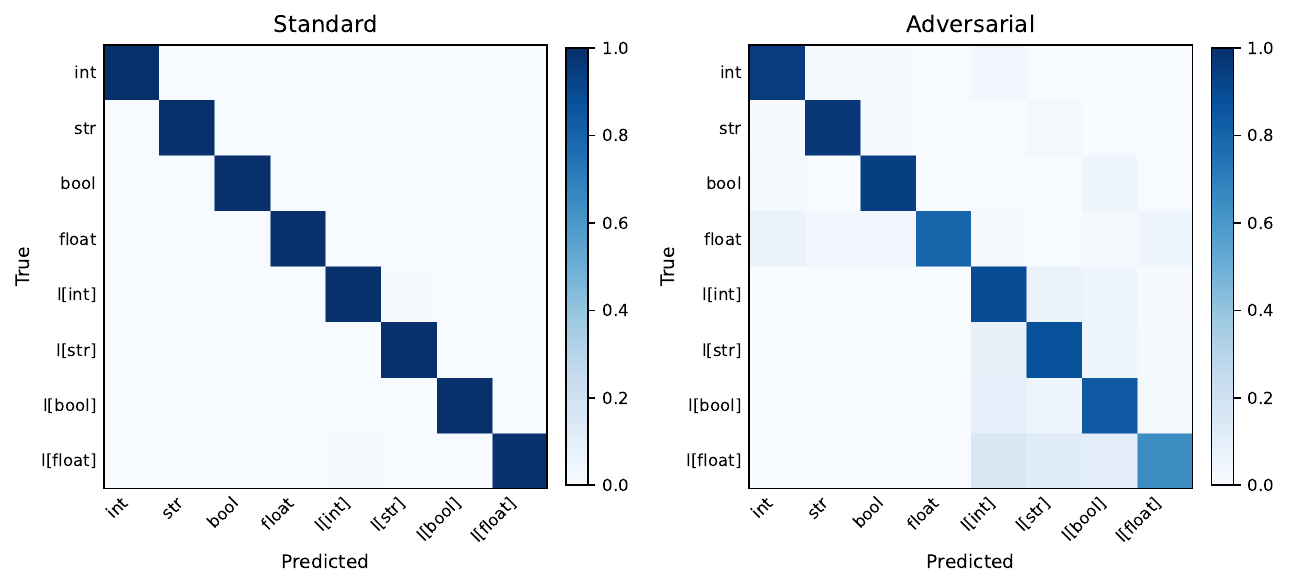}
    \caption{SantaCoder (1.1B) at layer 12.}
    \label{fig:cm-santacoder-task2-pyTag}
  \end{subfigure}
  
  \vspace{1.5em}
  
  \begin{subfigure}{\linewidth}
    \centering
    \includegraphics[width=\linewidth]{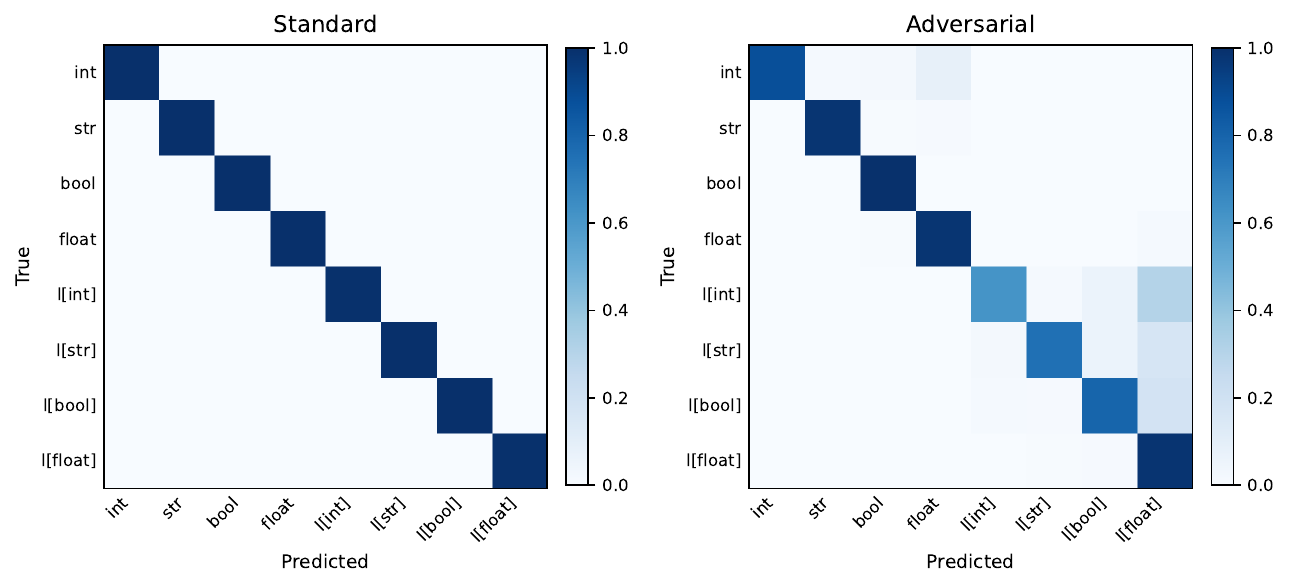}
    \caption{CodeLlama (7B) at layer 17.}
    \label{fig:cm-codellama-task2-pyTag}
  \end{subfigure}

  \caption{Row-normalized confusion matrices for Task 2 (Application Result). Probes are trained and evaluated on \texttt{pyTag}\ under standard (left) and adversarial (right) settings.}
  \label{fig:cm-task2-pyTag-combined}
\end{figure}

\begin{figure}[h]
  \centering
  \begin{subfigure}{\linewidth}
    \centering
    \includegraphics[width=\linewidth]{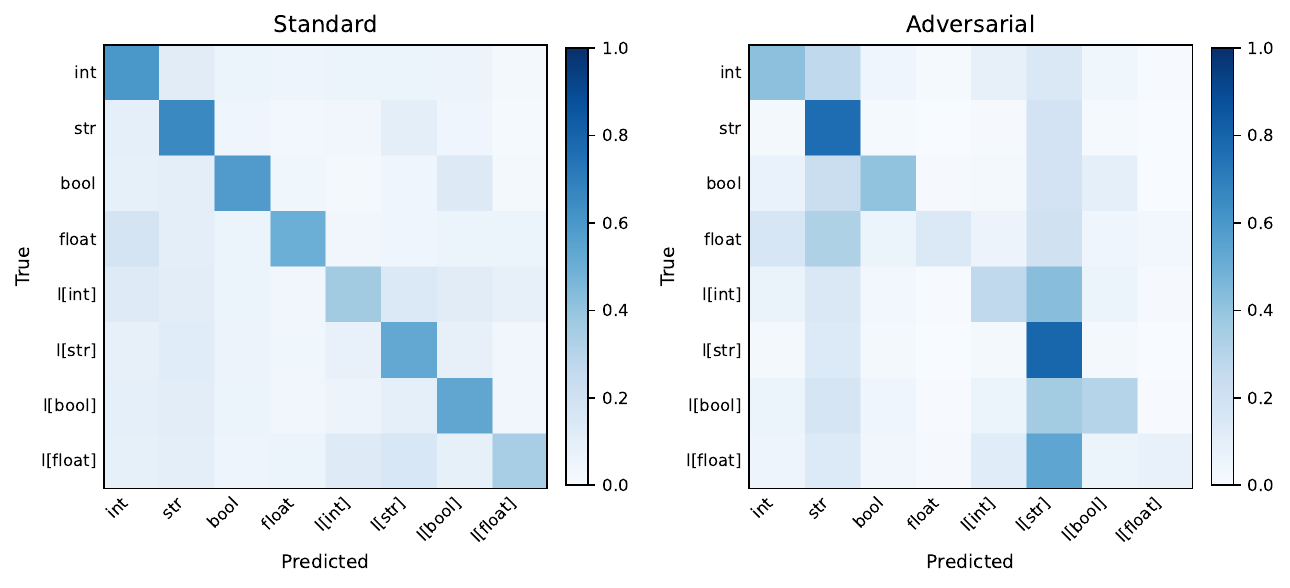}
    \caption{SantaCoder (1.1B) at layer 12.}
    \label{fig:cm-santacoder-task2-pyUnt}
  \end{subfigure}
  
  \vspace{1.5em}
  
  \begin{subfigure}{\linewidth}
    \centering
    \includegraphics[width=\linewidth]{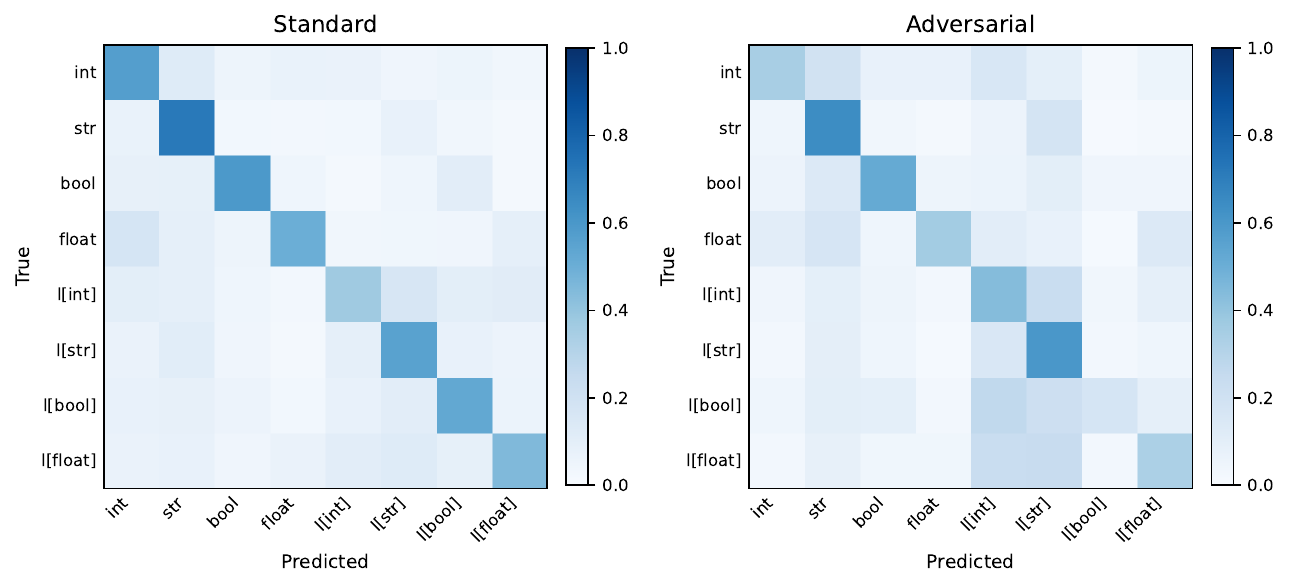}
    \caption{CodeLlama (7B) at layer 18.}
    \label{fig:cm-codellama-task2-pyUnt}
  \end{subfigure}

  \caption{Row-normalized confusion matrices for Task 2 (Application Result). Probes are trained and evaluated on \texttt{pyUnt}\ under standard (left) and adversarial (right) settings.}
  \label{fig:cm-task2-pyUnt-combined}
\end{figure}
\clearpage
\section{Layer-wise Selectivity (\Java, \pyTag)}
\begin{figure}[!ht]
  \centering
  \includegraphics[width=\textwidth]{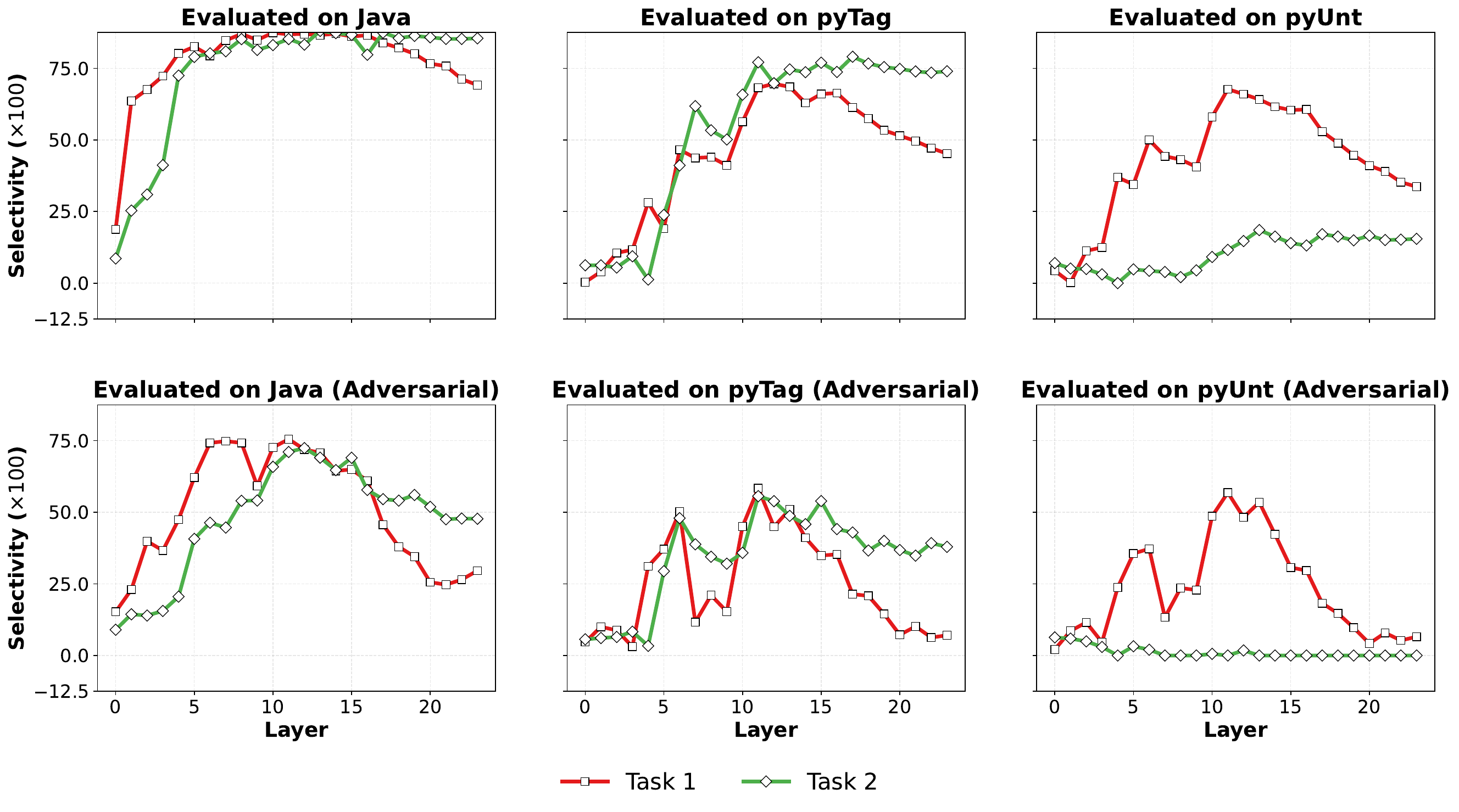}
  \vspace{0.2cm}
  \includegraphics[width=\textwidth]{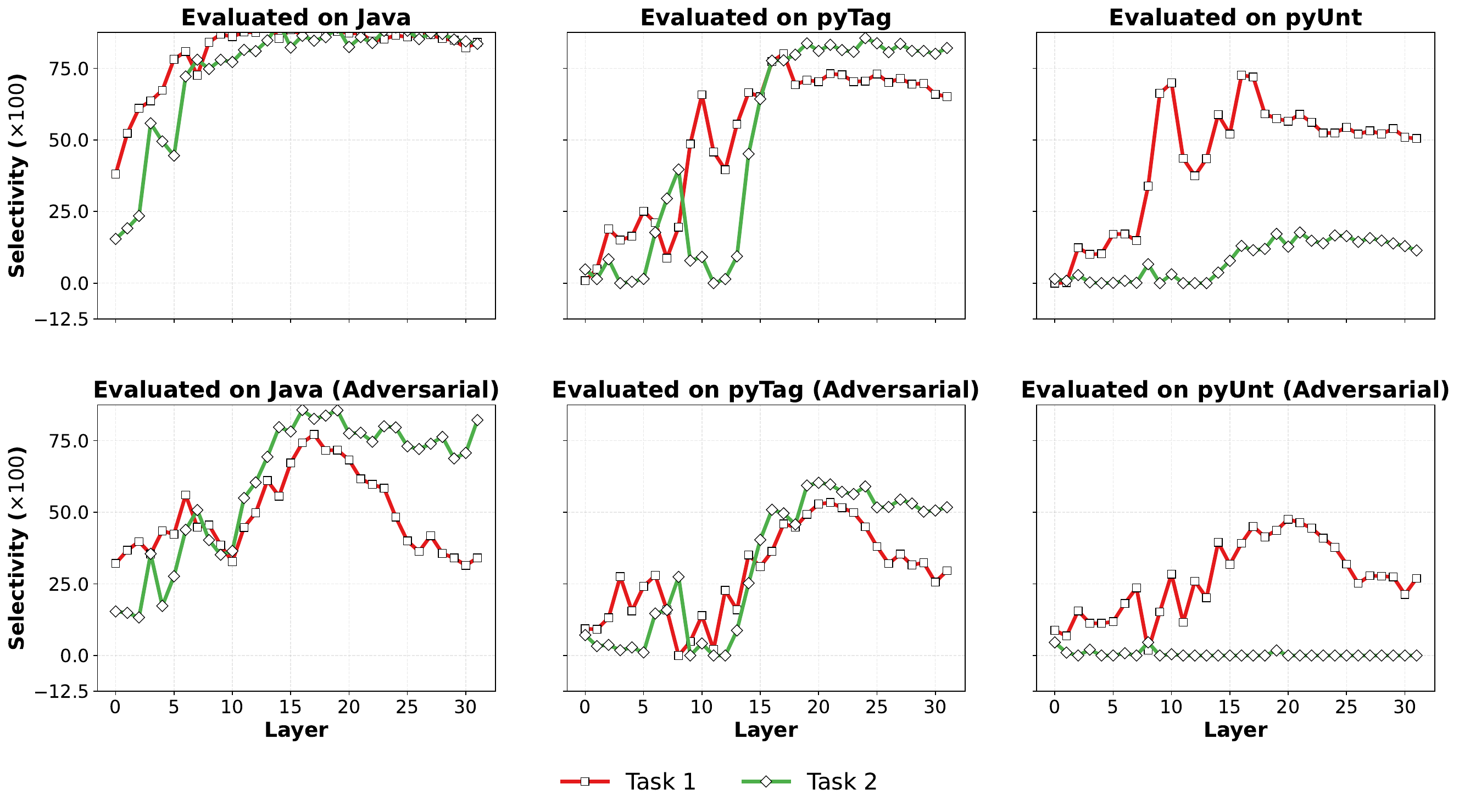}
  \caption{Layer-wise selectivity and accuracy (Tasks 1 \& 2) with probes trained on \Java and evaluated across all standard and adversarial partitions. \textbf{Top:} SantaCoder (1.1B). \textbf{Bottom:} CodeLlama (7B).}
  \label{fig:selectivity-combined-java}
\end{figure}
\begin{figure}[h]
  \centering
  \includegraphics[width=\textwidth]{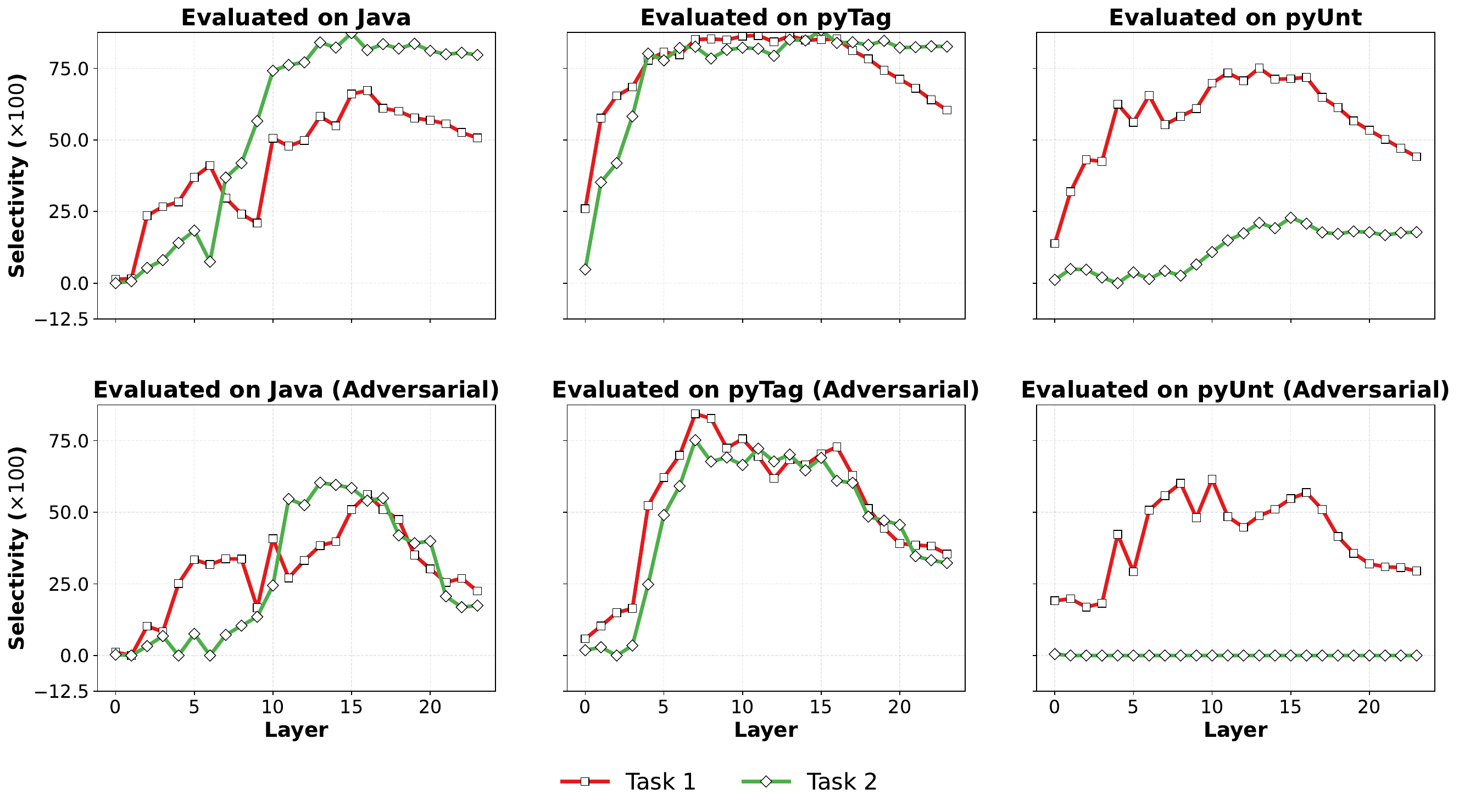}
  \vspace{0.2cm}
  \includegraphics[width=\textwidth]{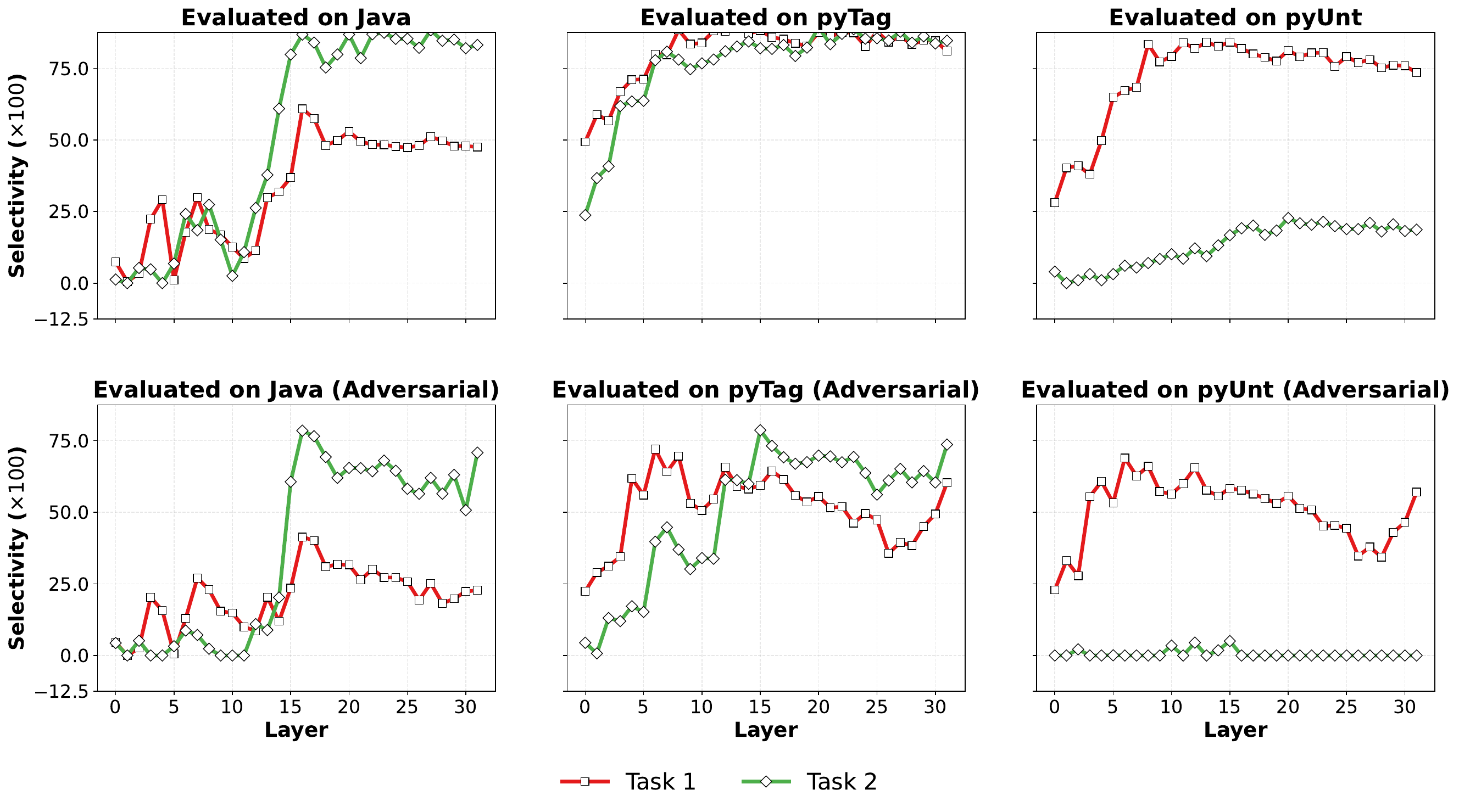}
  \caption{Layer-wise selectivity and accuracy (Tasks 1 \& 2) with probes trained on \pyTag and evaluated across all standard and adversarial partitions. \textbf{Top:} SantaCoder (1.1B). \textbf{Bottom:} CodeLlama (7B).}
  \label{fig:selectivity-combined-pyTag}
\end{figure}
\end{document}